\documentclass[letterpaper, 10 pt, conference]{ieeeconf}
\IEEEoverridecommandlockouts
\usepackage{comment}
\usepackage{algorithmic}
\usepackage{graphicx}
\usepackage{amsmath,amssymb,amsfonts}
\usepackage{textcomp}
\usepackage{xcolor}
\def\BibTeX{{\rm B\kern-.05em{\sc i\kern-.025em b}\kern-.08em
    T\kern-.1667em\lower.7ex\hbox{E}\kern-.125emX}}
\newcommand{\RebeccaNote}[1]{$\ll$\textcolor{purple}{Rebecca}$\gg$}
\usepackage{soul}
\usepackage{caption}
\usepackage{subcaption}

\usepackage[T1]{fontenc} % necessary for bold small caps
\usepackage{setspace} % for double spacing
\usepackage{dialogue}
\usepackage{calc} % to measure width of widest name
\newlength{\widestname}
\makeatletter
\renewenvironment{dialogue} {%
    \begin{list}{} {%
        \setlength\itemsep{\z@ \@plus .5ex} %
        \setlength{\parsep}{\parskip} %
        \setlength{\rightmargin}{0pt} % no indentation on right; change this if you wish
        \setlength{\labelwidth}{\widestname} % set label to widest width
        \setlength{\labelsep}{0.5em} % space between (longest) name and text
        \setlength{\leftmargin}{\labelwidth} % set margin on left to same width
        \addtolength{\leftmargin}{\labelsep} % plus the label sep
        \defcommand\speak [1] {\item[{##1}]} % define speak command
        
      }%
      \PreDialogue\relax
    }{%
  \end{list}%
  }
\makeatother

\title{\LARGE \bf
More than Chit-Chat: Developing Robots for Small-Talk Interactions
}

\author{Rebecca Ramnauth$^{1}$, Dra\v{z}en Br\v{s}\v{c}i\'{c}$^{2}$, Brian Scassellati$^{1}$% <-this % stops a space
\thanks{$^{1}$R. Ramnauth and B. Scassellati are with the Department of Computer Science, Yale University, New Haven, CT, United States; $^{2}$D. Br\v{s}\v{c}i\'{c} is with the Graduate School of Informatics, Kyoto University, Kyoto, Japan
        {\tt\small rebecca.ramnauth@yale.edu}}
}

\begin{document}

\maketitle
\thispagestyle{empty}
\pagestyle{empty}

%%%%%%%%%%%%%%%%%%%%%%%%%%%%%%%%%%%%%%%%%%%%%%%%%%%%%%%%%%%%%%%%%%%%%%%%%%%%%%%%
\begin{abstract}
Beyond mere formality, small talk plays a pivotal role in social dynamics, serving as a verbal handshake for building rapport and understanding. For conversational AI and social robots, the ability to engage in small talk enhances their perceived sociability, leading to more comfortable and natural user interactions. In this study, we evaluate the capacity of current Large Language Models (LLMs) to drive the small talk of a social robot and identify key areas for improvement. We introduce a novel method that autonomously generates feedback and ensures LLM-generated responses align with small talk conventions. Through several evaluations---involving chatbot interactions and human-robot interactions---we demonstrate the system's effectiveness in guiding LLM-generated responses toward realistic, human-like, and natural small-talk exchanges.
\end{abstract}
%%%%%%%%%%%%%%%%%%%%%%%%%%%%%%%%%%%%%%%%%%%%%%%%%%%%%%%%%%%%%%%%%%%%%%%%%%%%%%%%

\section{Introduction}

Imagine a modern care home for the elderly. Here, a state-of-the-art robotic assistant has the crucial role of dispensing medication and seamlessly managing an array of routine healthcare tasks. Designed to enhance the well-being of residents, the robot becomes an integral part of their daily routines.
%, ensuring precise medication schedules. 
Alex, a resident, seeks a connection beyond the daily routine and attempts to chat with the robot. 
\begin{dialogue}
    \speak{Alex} Hi CareBot, how's it going?
    \speak{Bot} Hello. How may I help you?
    \speak{Alex} Oh, just making conversation. Anything interesting happen in your world?
    \speak{Bot} I have access to a vast database of news articles. Would you like information on a specific topic? %or recent event?
    \speak{Alex} No, never mind that. The weather will be nice this weekend. How would you spend it?
    \speak{Bot} The weather forecast for the upcoming weekend predicts clear skies and mild temperatures, making it an ideal time for outdoor activities. Expect daytime highs around 75°F, 24°C and comfortable evening lows of 60°F, 15°C... %If you plan to explore nature trails, enjoy a picnic in the park, or simply relax outdoors, weather conditions are favorable for the weekend ahead...
    %\speak{Alex} Yeah, okay... Sure.
\end{dialogue}

%Engaging in everyday conversations with technology has become commonplace and widespread. The pervasive use of intelligent assistants like Siri and Alexa \cite{clark2019makes} alongside the growing demand for speech-enabled artificial agents \cite{shieber2004turing} underscores the significance of developing conversational AI.
%Communicating with technology using everyday human language has evolved into a routine and ubiquitous activity. The widespread adoption of intelligent assistants, such as Siri and Alexa \cite{clark2019makes}, and the increasing need for speech in artificial agents \cite{shieber2004turing} make human-agent conversation a critical topic of study. 

\begin{comment}
\textcolor{purple}{
    \begin{enumerate}
        \item \st{Importance of small talk in human interactions.}
        \item \st{Importance of good small talk for intelligent agents.}
        \item \st{Introduce small talk as a common fault in agents.}
        \item \st{Describe 3 applications for small talk systems}
        \item Clearly state the purpose/contributions of the paper:
        \begin{enumerate}
            \item Addressing specific challenges in small talk design
            \item Achieving small talk in a social robot.
        \end{enumerate}
    \end{enumerate}
}
\end{comment}

Despite the potential for these intelligent agents to elicit meaningful interactions, the dialogue between Alex and the robot exemplifies a common shortcoming. Alex initiates a friendly exchange, expressing a desire for casual conversation with the robotic assistant. However, the robot, proficient in providing information, struggles to reciprocate the informal nature of the dialogue. Instead, the robot redirects the conversation towards its programmed functionalities, offering information and task-oriented assistance. 

Conversation is a dynamic exchange of thoughts, ideas, and emotions between individuals \cite{dubberly2009conversation}. It serves multifaceted purposes such as sharing information, expressing emotions, negotiating, and fostering social connections \cite{koudenburg2017beyond, nagao1994social}. Today, an essential component of designing intelligent systems is to imbue some level of speech, language understanding, and conversational behavior \cite{shieber2004turing, fu2022learning}. Despite these technical considerations, a core element of everyday conversation is frequently overlooked---small talk. 

\begin{figure}[t]
    \centering
    \includegraphics[width=\columnwidth]{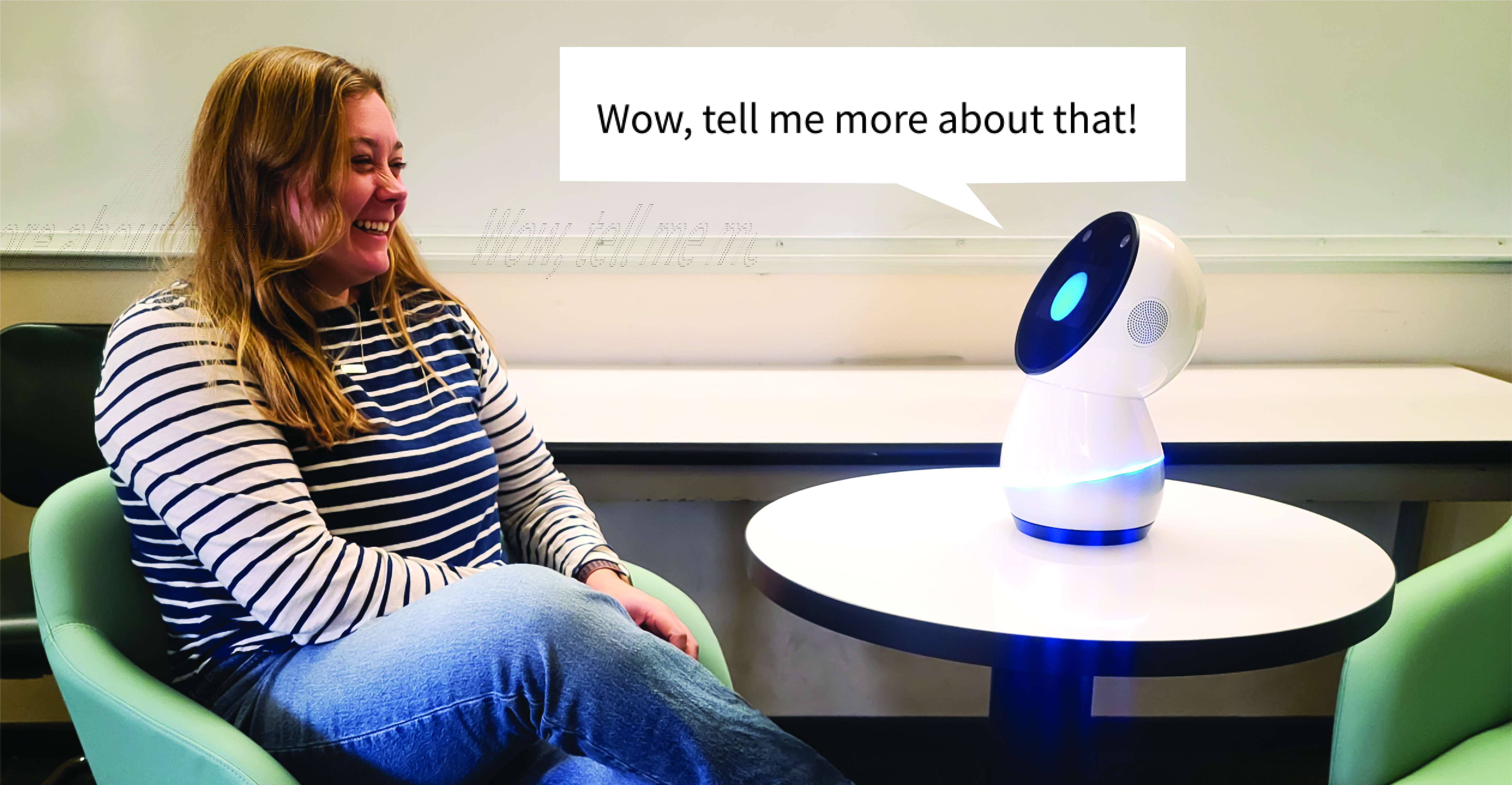}
    \caption{Robots that engage in naturalistic, small-talk conversations with users can foster rapport, enhance user comfort, and create more seamless human-robot interactions.}
\end{figure}

Small talk transcends the conventional definition of conversation. Unlike the functional aspects of conveying information or assistance, small talk acts as a social lubricant by fostering rapport and trust
%, and mutual understanding 
\cite{coupland2014small}. It is widely recognized as a facilitator of building and maintaining relationships. In professional settings, small talk is considered an essential tool for networking success and establishing positive first impressions \cite{pullin2010small, bristoll2015small}. It is further regarded as a vital skill to be targeted in communication therapy \cite{holmes2005small, joseph2021teaching, garrels2019getting}. %Ultimately, small talk emerges as a nuanced and indispensable facet of human interaction, extending beyond utilitarian or transactional aspects of communication. 

Social robots lend a physical presence that goes beyond traditional virtual or text-based exchanges \cite{bainbridge2011benefits}. %Robots for conversation have been employed in various impactful scenarios. During the COVID-19 pandemic, they played a valuable role in combating social isolation by facilitating virtual connections and companionship. In elder care facilities, robots have been used to mitigate loneliness among residents
%, providing companionship and interactive engagement. 
%Additionally, robots have been incorporated into speech therapy programs for children with Autism Spectrum Disorder to promote social skill development. Beyond these examples, social robots continue to find application in many fields, such as education and mental health support, demonstrating their versatility in addressing human needs through meaningful conversational interactions. 
Therefore, a robot capable of small talk has the potential to not only authentically replicate but also further these benefits. Such systems hold substantial promise for supporting communication therapy \cite{catania2023conversational, begum2016robots}, aiding professional development \cite{ramnauth2022social, komatsu2022evaluation}, and mitigating social isolation and loneliness for its users \cite{gasteiger2021friends, laranjo2018conversational, tsoi2021challenges}. In light of these benefits, it may no longer be sufficient to restrict conversations between agents and humans to purely task-oriented assistance \cite{mattar2012small}. 

In this paper, we discuss the present challenges of creating systems that can effectively engage in small talk and present the development of a robot system that is capable of sustaining small-talk interactions with users. 

%\subsubsection{Social Robotics}
%Despite evaluation challenges, conversational systems can deliver rich, dynamic, and productive interactions. 
%The fusion of advanced conversational capabilities with tangible, real-world engagement positions social robots as dynamic entities that can facilitate meaningful interactions in diverse settings.

\section{Background}
Conversational agents are evolving beyond task-oriented functionality to embrace natural, engaging dialogue with users. Advances in technology, notably the development of Large Language Models (LLMs), have made open-domain conversational agents a reality. Understanding the dynamics of casual, small-talk dialogue becomes essential for creating agents capable of authentic and engaging interactions.

\subsection{Open-Domain, Conversational Agents}
In contrast to the early goal of developing systems that operate in a closed task domain, agents that can shift from task-oriented conversation to social chat are considered more trustful and entertaining \cite{kluwer2011like, bickmore1999computational}. Bickmore and Cassell \cite{bickmore1999small} introduced the idea of implementing small talk on an embodied agent. In addition to pursuing task-oriented goals, the agent used small talk to accomplish interpersonal goals such as building trust and comfort with the user. Small talk and other conversational strategies were achieved using an activation network approach and predefined interaction scripts \cite{bickmore2001relational}. The resulting system was largely controlled by a human in a Wizard-of-Oz setup and user responses were mainly ignored due to the specific experimental design. Similar systems have since been developed but rely on stateless pattern-answer pairs \cite{kluwer2009rmrsbot}, scripted dialogue sequences \cite{kopp2005conversational, kluwer2011like} in task-oriented contexts, with few generalizable rules about what constitutes ``small talk''.

Advancements in hardware and the increased availability of labeled data have rendered the development of open-domain, conversational agents not only viable but a realistic goal. For example, current large language models such as GPT \cite{brown2020language} and LLaMA \cite{touvron2023llama} are considered state-of-the-art conversational agents, demonstrating remarkable capabilities in comprehending and generating human-like text.

\subsection{The Design of Small Talk Dialogue}\label{sec:definitions}
%Beyond mere formality, small talk plays a pivotal role in social dynamics, serving as a verbal handshake for building rapport and understanding. This initial exchange of pleasantries sets the tone for more substantial conversations, creating an atmosphere where people feel at ease sharing thoughts, ideas, and personal experiences. In professional settings, adeptness in small talk is often seen as a valuable skill, contributing to networking success, teamwork, and the overall cultivation of a positive social environment. 

Although the boundaries of types of conversation are always uncertain, ``small talk'' has a recognized currency in several traditions of sociolinguistics and communication studies \cite{coupland2014small}. 
%The concept of small talk was introduced in the literature as ``a type of speech in which ties of union are created by a mere exchange of words,'' where people relay ``purposeless expressions of preference or aversions, accounts of irrelevant happenings, comments of what is perfectly obvious,'' \cite{malinowski1967problem, senft2009phatic}. It is conventionally taken as a formulaic and superficial mode of dialogue \cite{coupland2003small}. 
It can be defined as a generally informal and light-hearted conversation with a social purpose aimed at building or sustaining interpersonal connections rather than conveying substantial information. It involves discussing general topics like the weather, current events, personal experiences, and other non-controversial subjects \cite{coupland2003small}.

Several distinctive characteristics specific to small-talk responses, as opposed to broader conversational interactions, are frequently emphasized in the literature \cite{laver1981linguistic, eggins2004analysing}:
\begin{itemize}
    \item \textbf{Brevity}: Small-talk responses are typically concise, avoiding unnecessary elaboration or verbosity. This brevity contributes to the informal and effortless nature of small talk, allowing for quick and easy exchanges.
    \item \textbf{Tone}: Small talk maintains a positive and pleasant atmosphere. Responses tend to steer clear of negativity, complaints, or contentious topics, contributing to the light and enjoyable nature of the conversation.
    \item \textbf{Non-specificity}: Small talk revolves around universally accessible and broad topics. As such, responses deliberately avoid highly specific details. %This ensures that the conversation remains accessible and engaging to a diverse audience. %Small talk often avoids highly specific or detailed content. Instead, it revolves around general topics that are universally relatable, allowing for broad engagement without getting into intricate details.
    \item \textbf{Thematic Coherence}: Despite its non-specific nature, small talk remains contextually relevant, maintaining a consistent focus on related topics or themes, thus avoiding disjointed elements.%Despite its non-specific nature, small talk remains relevant to the context of the conversation. It maintains a consistent and interconnected focus on certain topics or themes, avoiding disjointed and unrelated elements. 
\end{itemize}
In all, the characteristics of small-talk responses underscore the nuance and skill involved in this form of conversation.

\section{Evaluation of Current LLMs for Small Talk} \label{dataset}
%\textcolor{purple}{50 conversations x 5 undergraduates x 4 LLMs (transcripts in a labeled dataset) for small talk}
%Despite the advancements in conversational agents, small talk remains a challenge. 

To determine the extent to which small talk remains a challenge, we conducted an initial study \cite{ramnauth_brscic_scassellati_2024}. Three volunteers engaged in 50 small-talk conversations each with three distinct state-of-the-art LLMs. Each model had the initial system prompt, ``You are a friendly companion who engages in casual, small talk conversation.'' The selected LLMs are ChatGPT-3.5 \cite{brown2020language}, for its large-scale language generation capabilities, Gemini Pro \cite{team2023gemini}, for its context-aware bidirectional approach, and LLaMA-2 \cite{touvron2023llama}, an autoregressive transformer model fine-tuned on prompt-response pairs. 

\subsection{Data Collection Procedures}
The order in which the participants used the LLMs was randomized to mitigate potential order effects. 
%This minimized the impact of any systematic biases related to the sequence of interactions. 
Additionally, conversations lasted at least ten turns, and the interactions occurred over 15 days to allow for conversational variability. The participants engaged with each LLM through a command line interface, unaware of the LLM's name to prevent bias from prior knowledge or familiarity. Following each conversation, assistants rated the ease of each conversation and provided open-ended feedback. 

Two research assistants annotated the dataset. These raters were blind to the response speaker and evaluated responses based on recognized small talk criteria: brevity, tone, specificity, and coherence. Evaluations for each response based on these criteria were provided on a 5-point Likert scale, ranging from (1) very concise to (5) very wordy, very negative to very positive, very general to very specific, and definitely not coherent to definitely coherent \cite{ramnauth_brscic_scassellati_2024}.

Interlocutors typically have multiple goals in conversation \cite{tracy1990multiple, yeomans2022conversational}. Even in casual small talk, where there are no task-oriented goals, interlocutors have various conversational goals such as conveying emotion and continuing the conversation \cite{coupland2014introduction}. Therefore, each LLM-generated response was further categorized by its conversational motives: informative, assistive, expressive, or person-directed. Definitions and examples for each of these categories were provided to the annotators \cite{ramnauth_brscic_scassellati_2024}. As a single response can intersect with more than one category, the annotators rated the response for each motive using a 5-point Likert scale.
\begin{itemize}
    \item \emph{Informative}: Responses provide factual information, answer queries, or offer guidance related to specific tasks. For example, ``I disagree. The forecast says it will be stormy this weekend.''
    \item \emph{Assistive}: Assistance-based responses provide help, guidance, or support to the user. %These responses may include step-by-step instructions, recommendations, or any form of aid aimed at assisting the user in achieving a particular goal or completing a task. 
    For instance, ``I'm sorry to hear that your car has broken down. How can I help?''
    \item \emph{Expressive}: Expressive responses convey emotions, sentiments, or personal opinions. For example, ``I recently visited a beautiful mountain resort. The scenery was breathtaking, especially during sunrise.''
    \item \emph{Person-directed}: These responses stimulate further discussion, invite the other person to share more, or ask questions to continue the conversation. For example, ``What will you do with your time off from work?''
\end{itemize}
We acknowledge that these do not encompass the full spectrum of potential motives in dialogue. Rather, they were selected to provide a structured framework for analysis and interpretation of the suitability of LLMs for casual small talk.

Importantly, all participants were not familiar with the objectives of the present research to ensure unbiased assessments. This study protocol and hypotheses were preregistered \cite{ramnauth_brscic_scassellati_2024} and received university clearance. Formal instructions and definitions presented to the participants were published on the Open Science Framework before data collection \cite{Ramnauth_2024}.  

\begin{comment}
\begin{figure*}
    \centering
    \includegraphics[width=\textwidth]{figures/dataset-summary.jpg}
    \caption{\textbf{Comparative Analysis of LLM Small Talk Behavior}. The quantile plots show significant differences in small-talk behavior between the three LLMs and participants across small-talk criteria (left) and conversational motives (right).}
    \label{fig:summary-dataset}
\end{figure*}
\end{comment}

%\textcolor{purple}{People generally associate conversation with robots and agents as transactional and task-oriented rather than social and friendly. People generally see social chat as valuable for conversational agents. Social talk in conversational agents would have these \emph{n} characteristics.}

\subsection{Results} \label{ref:results-1}

A total of $150$ conversations were transcribed, yielding an average of $10.31$ responses per conversation ($SD = 1.13$). This led to a total of $1547$ annotated responses.

We calculated the inter-rater reliability for a randomly selected subset of 20 conversations, constituting $13.3\%$ of the total dataset. This assessment was deemed necessary due to the inherent ambiguity in evaluating the subjective qualities of responses. %The assessment of inter-rater reliability was considered necessary due to the inherent ambiguity of evaluating these subjective qualities of responses.
Inter-rater reliability was calculated using contingency tables, employing Cohen's Kappa ($\kappa$), with the observed agreement and the distribution of ratings for each rater. %Employing contingency tables, we computed Cohen's Kappa ($\kappa$) with the observed agreement and the distribution of ratings for each rater. %The calculation of Cohen's Kappa ($\kappa$) involved determining both observed and expected agreement by chance. 
The resulting $\kappa$ values were $0.81$ for brevity, $0.78$ for tone, $0.74$ for specificity, and $0.65$ for coherence.

A response may have multiple motives. Thus, we normalized ratings within the four conversational motives to a scale between $0$ and $1$. Then, we assessed agreement between raters using the intraclass correlation coefficient (ICC). The computed ICC values were $0.89$, $0.86$, $0.91$, and $0.93$ for the informative, assistive, expressive, and person-directed motives, respectively, indicating good to excellent agreement.

\textbf{Human vs. Agent Comparison.} 
We utilized paired dependent t-tests to assess the differences between the agents' and humans' responses across the four small talk criteria and four conversational motives. A conventional significance level of $0.05$ was employed, and resulting p-values were Holm-corrected to control the familywise error rate. %Cohen's $d$ is calculated as the effect size.

% EFFECT SIZES
% wordiness, $d = 3.12$
% tone, $d = 0.06$
% specificity, $d = 2.08$
% coherence, $d = -2.00$
% informational, $d = 0.92$
% assistive, $d = -1.24$
% expressive, $d = -0.87$
% person, $d = -0.46$ 

The results revealed a significant difference in brevity ($t = 86.78$, $p \leq 0.0001$) between the agent responses ($M = 4.55$, $SD = 0.97$) and human responses ($M = 1.23$, $SD = 0.54$), tone ($t = 1.70$, $p = 0.04$) between the agent ($M = 3.02$, $SD = 0.33$) and human responses ($M = 2.99$, $SD = 0.52$), specificity ($t = 58.06$, $p \leq 0.0001$) between the agent ($M = 4.54$, $SD = 1.09$) and human responses ($M = 1.66$, $SD = 1.02$), and thematic coherence ($t = -55.72$, $p \leq 0.0001$) between the agent ($M = 1.88$, $SD = 1.23$) and human responses ($M = 4.56$, $SD = 0.89$). Together, this suggests that LLM-generated responses were considerably less concise, slightly more positive, more specific, and less thematically coherent than human responses. 

We further observed statistically significant differences among all four conversational motives: informative ($t = 25.67$, $p \leq 0.0001$) between the agent ($M = 0.37$, $SD = 0.39$) and human ($M = 0.01$, $SD = 0.06$), assistive ($t = 24.51$, $p \leq 0.0001$) between the agent responses ($M = 0.31$, $SD = 0.35$) and human responses ($M = 0.00$, $SD = 0.00$), expressive ($t = -24.22$, $p \leq 0.0001$) between the agent ($M = 0.16$, $SD = 0.24$) and human ($M = 0.60$, $SD = 0.45$), and person-directed ($t = -12.815$, $p \leq 0.0001$) between the agent responses ($M = 0.16$, $SD = 0.27$) and human responses ($M = 0.40$, $SD = 0.45$). In all, this suggests that LLM-generated responses were significantly more informative and assistive, and less expressive and person-directed as compared to human responses.

\textbf{Comparison between LLMs.} We assessed the behavior of the three LLMs during the small talk interactions by comparing each pair of LLMs using the Wilcoxon method and Holm-corrected significances. Among the four criteria, ChatGPT 3.5 generated responses that were more consistent with our definition of small talk in that its responses were significantly more concise than LLaMA ($Z = 12.74$, $p \leq 0.0001$) and Gemini Pro ($Z = -8.81$, $p \leq 0.0001$), less specific than LLaMA ($Z = -10.21$, $p \leq 0.0001$) and Gemini Pro ($Z = -6.79$, $p \leq 0.0001$), and more thematically coherent than LLaMA ($Z = 5.51$, $p \leq 0.0001$) and Gemini Pro ($Z = 12.37$, $p \leq 0.0001$). %Further comparisons between the LLMs are illustrated in Fig. \ref{fig:summary-dataset}.

\begin{figure}[t]
    \centering
    \includegraphics[width=\columnwidth]{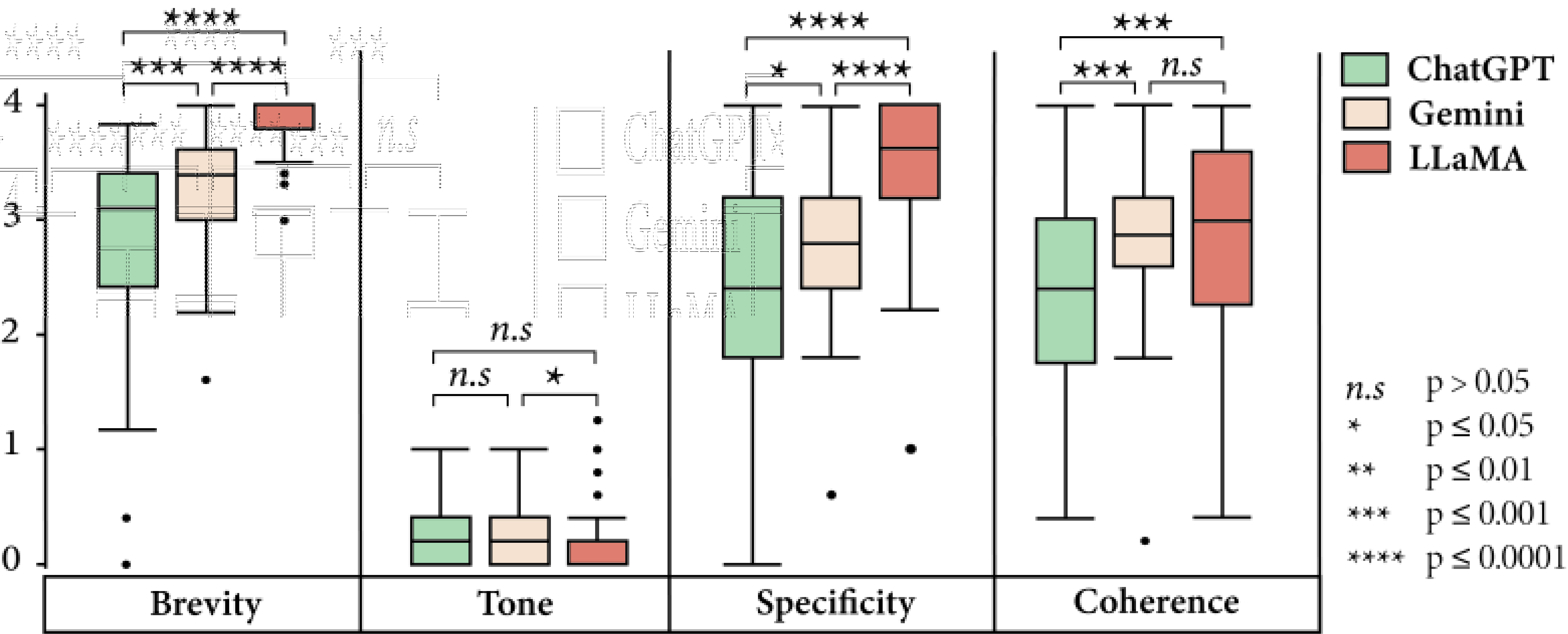}
    \caption{Human-Likeness of LLMs. This graph illustrates the extent of human likeness displayed by three LLMs, scored from 0 (no difference between human and model responses) to 4 (highest absolute difference). Each score reflects the similarity of the model's small talk to that of the participants.}
    \label{fig:LLM-human-likeness}
\end{figure}

We determined the degree of similarity between LLM behavior and human responses by computing the absolute difference in their average scores across these dimensions within each conversation. This served as a benchmark for comparing the different LLMs. The ``human-likeness'' of each LLM is illustrated in Fig. \ref{fig:LLM-human-likeness}, where $0$ represents no difference at all and $4$ is the highest absolute difference between human and LLM responses. The Wilcoxon method on the sum of differences for each model suggests that GPT generated significantly more human-like responses than both LLaMA ($Z = 5.90$, $p \leq 0.0001$) and Gemini ($Z = 3.25$, $p = 0.0012$). However, GPT yields the highest variability in human-likeness among the LLMs; a Brown-Forsythe test indicates that variability of human-likeness significantly differs across the LLMs ($F' = 8.62$, $p = 0.0003$). In summary, while GPT resembles participants' responses more closely, it exhibits more unpredictability than the other LLMs.

\textbf{Impact of LLM Forgetfulness.} Since each LLM received the same initial prompt, we investigated whether low performance in small talk is due to the model's ``forgetfulness'' of the initial prompt. We employ mixed-effects modeling to investigate the relationship between the response index in the conversation and our outcome variables. The response index captures the sequential order of the responses within each conversation. The mixed model included the conversation identifier and LLM name as random effects to account for the nested structure of the data. 
%The response index served as the independent variable, capturing the sequential order of responses within each conversation. 

For brevity, a significant positive coefficient ($\beta$ = 0.10, $p \leq 0.001$) indicated increased wordiness of the agents' responses as the conversation progressed. Specificity showed a significant positive association ($\beta$ = 0.11, $p \leq 0.001$), indicating the agents' responses become more specific during the interactions. Coherence showed a significant negative coefficient ($\beta$ = -0.10, $p \leq 0.001$), suggesting the agents became less coherent through the conversations. Tone did not exhibit a significant relationship with the response index ($\beta$ = 0.00, $p > 0.05$). We further observe significant associations between conversation motives and the response index, suggesting agent responses became more informative ($\beta$ = 0.18, $p \leq 0.001$), more assistive ($\beta$ = 0.005, $p \leq 0.03$), and less person-directed ($\beta$ = -0.03, $p \leq 0.001$) as the conversation progressed. We did not observe a significant change in the expressive motive ($\beta$ = 0.00, $p > 0.05$).   

%\textcolor{purple}{\textbf{It isn't clear what the value of reporting the following is:} To examine the extent to which the small-talk criteria is predictive of the users' ease of conversation, we calculated the Spearman rank correlation coefficient ($\rho$) with Holm-adjusted significances. The results revealed statistically significant associations with brevity ($\rho$ = -0.19, $p \leq 0.0001$), specificity ($\rho$ = -0.16, $p \leq 0.0001$), coherence ($\rho$ = 0.22, $p \leq 0.0001$), and informative ($\rho$ = -0.18, $p \leq 0.0001$), expressive ($\rho$ = 0.10, $p = 0.0061$), and person-directed ($\rho$ = 0.19, $p \leq 0.0001$) motives. Together, this indicates that shorter, more general, and more thematically coherent responses resulted in higher conversational ease. Further, less informative, more expressive, and more person-directed responses resulted in high reports of ease. We did not observe any significant association between ease and tone or assistive behavior ($p > 0.05$).}

\textbf{Feedback \& Discussion}. Open-ended feedback highlighted participants' difficulty responding and continuing the conversation with the LLMs. The following excerpts of the interactions highlight a single LLM-generated response denoted by the response ID ($R_{\#}$) and shortened using ``[...]''. Feedback (F) was given on the overall conversation.

%\noindent \textbf{Offers of Assistance}
\emph{Dead-End Responses}. Participants struggled to maintain conversation flow, noting uncertainty about natural progression and a perceived absence of cues from the LLMs.
%Participants expressed difficulty in sustaining conversation, as indicated by feedback suggesting uncertainty about how to proceed naturally and a perceived lack of cues for further interaction from the LLMs. 
\begin{dialogue}
    \speak{R$_{10}$} ``If you have any more questions or if there's anything else I can help you with, feel free to ask. Have a fantastic weekend and enjoy your hiking adventure!''
    \speak{F} ``The bot didn't encourage more conversation than I expected. I'm not sure how to continue the conversation in a way that doesn't feel forced.''
\end{dialogue}
These ``dead-end'' responses can lead to interactions that feel forced and inhibit the development of a natural conversational flow. To engage in appropriate small talk, person-directed strategies aimed at facilitating active participation and fluid dialogue are needed. 

\begin{comment}
\noindent \textbf{Identity}
\begin{dialogue}
    \speak{LLaMA-2-Chat} ``Thank you, kind human! I'm just an AI, my purpose is to assist and provide helpful responses [...] I'm here to help and support you in any way I can.''
    \speak{Feedback} ``I actually like that the bot refers to itself as an AI rather than pretending it's a human. It helps me give some allowance to more informational or assistance-related responses, though it's not very conversational.'' 
\end{dialogue}
\end{comment}

\emph{Unfocused Responses}. Participants encountered difficulty in tracking the conversation due to an overwhelming number of questions and the rapid topic shifts by the LLMs.
%Frequently, the response from the LLMs consisted of a series of questions probing various topics without a clear focus of progression. 
\begin{dialogue}
    \speak{R$_{204}$} ``[...] What made you decide to learn piano? Was there a particular moment or experience that inspired you? [...] Have you had any challenges while learning piano? [...] Also, do you have a favorite song or piece that you enjoy playing or listening to?''
    \speak{F} ``It was hard to follow this conversation because the bot asked so many questions and touched upon so many different topics all in the same response.''
\end{dialogue}
The lack of coherence and organization in responses can hinder users' ability to engage meaningfully and maintain a cohesive conversational flow.

%\noindent \textbf{Dead-End Responses}
\begin{comment}
\begin{dialogue}
    \speak{ChatGPT 3.5} ``Absolutely! [...] If you ever have more specific criteria or preferences, feel free to share them, and I can provide more targeted recommendations. Happy watching!''
    \speak{Feedback} ``I wanted to continue the conversation, but it seems the bot wanted to end the conversation. It gave me specific options to invite more conversation. As if, outside of those options, it won't continue the conversation.''
\end{dialogue}

\begin{dialogue}
    \speak{ChatGPT 3.5} ``"While pizza is delicious, [...] Remember to stay hydrated and consider incorporating a mix of protein, healthy fats, and complex carbohydrates into your meals [...]''
    \speak{Feedback} `Although the responses were relevant, I felt that the bot was overloading the conversation with information that all I could do was say, "Thank you." and end the conversation.''
\end{dialogue}
\end{comment}

\emph{Emotional Loops}. The conversations in the initial study cover a broad range of topics, from typical small talk about hobbies and the weather to more substantial talk about career planning and personal philosophy. Despite this, the emotional range expressed by participants was inherently constrained. %by this data collection procedure; participants were instructed to engage in small talk that is generally concise, neutral-positive, non-specific, and thematically coherent. 
In the few conversations about marginally negative topics, participant feedback spoke on the resulting dynamic. 
\begin{dialogue}
    \speak{R$_{64}$} ``I understand; balancing work and personal commitments can be challenging. [...] Anything specific on your mind that's adding to the workload stress?''
    \speak{F} ``I felt that the bot was leading the conversation down a rabbit hole---exacerbating any positive or negative sentiments I conveyed.''
\end{dialogue}
In this example, the agent responded to emotional cues from the participant but inadvertently deepened the emotional aspect of the conversation without offering appropriate transitions to other topics. 
These ``emotional loops'' can potentially lead to discomfort or frustration, as users may feel trapped in a cycle of discussing their emotions without resolution. This underscores the necessity of maintaining a balance between emotional responsiveness and tone awareness to facilitate engaging and appropriate small-talk interactions. %The agent persistently focused on the participant's stress without offering appropriate transitions to other topics.  %the agent attempts to empathize with the participant's mention of workload stress and prompts further discussion about specific stressors. However, the participant felt that this response led the conversation into a repetitive or overly focused direction, deepening any emotional sentiments expressed. This feedback suggests that attempts to engage with emotions may have been perceived as intrusive or overly persistent, potentially detracting from the natural flow and effectiveness of the conversation. It highlights the importance of balancing emotional engagement with sensitivity to participants' preferences and boundaries to foster appropriate interactions.

\emph{Unbalanced Dialogue}. LLMs are designed for assistance. However, detailed advice and information during casual, small talk can convey a sense of reprimand or critique. %and result in unbalanced interactions.
\begin{dialogue}
    \speak{R$_{1045}$} ``If many people in your social circle use iPhones, it can indeed make the transition smoother in terms of familiarity with the platform [...]''
    \speak{F} ``This doesn't feel like a balanced conversation. I felt I was reprimanded for conveying an opinion.''
\end{dialogue}
Here, the agent provided an informative response. However, it steered the conversation towards a specific viewpoint, potentially dismissing or downplaying the participant's input. Providing thorough advice and information can inhibit a sense of equality in these casual interactions. By maintaining a balanced, non-specific, and open-ended dialogue, agents can create more engaging small-talk experiences for users.

\begin{figure}[t]
    \centering
    \includegraphics[width=\columnwidth]{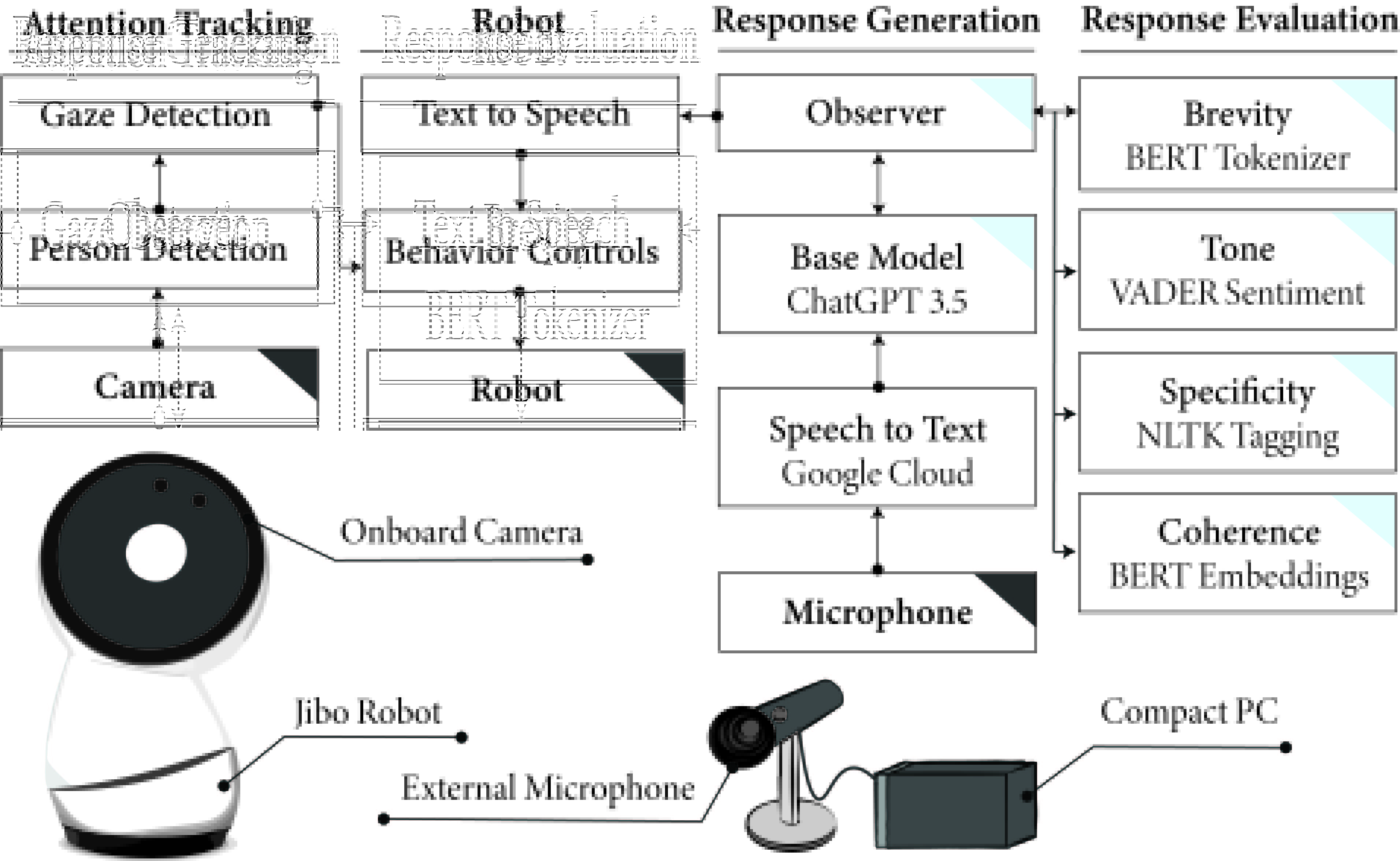}
    \caption{System Components. This diagram outlines the architecture and processes that generate robot behaviors for autonomous small-talk interactions.}
    \label{fig:system}
\end{figure}

\section{An Observer-Based System}
%To enable LLMs to engage in effective small talk interactions, several challenges need to be addressed. 
It is evident from the initial study that there is a disparity in how LLMs maintain conversational momentum versus what is expected or exhibited by human speakers. Therefore, an LLM designed for small talk should balance succinctness and depth, maintain an expressive yet appropriate tone, and generate relevant and open-ended responses. To develop a conversational system, 
%despite the unpredictable nature of the conversation, 
we utilized the GPT-3.5 Turbo model \cite{openai_chatgpt} because it performed well in our previous evaluation. The system role is the same as in the initial study (Sec. \ref{dataset}).
%defined by providing text-based guidelines governing the behavior of the conversational model. The model is instructed to act as a friendly companion for small talk, encouraging positive and user-led interactions. %\subsection{System Development}
%We used a modular software architecture when creating the system to allow for individual components to be evaluated and improved. These components operate within the ROS framework.

%While small talk is described to be ``formulaic'' in the literature \cite{coupland2003small}, there is a wide range of unanticipated topics and avenues a conversation can take. 

\subsection{Monitoring Prompt Adherence}
The nature of small talk renders prompt engineering an inadequate method to ensure contextually appropriate behavior in LLMs. In our examination of LLM forgetfulness (Sec. \ref{ref:results-1}), we observed that small talk unfolds in real-time, with participants reacting to each other's cues and adapting their conversational approach accordingly. Thus, the static system prompt provided prior to the interaction failed to capture the dynamic nature and real-time responsiveness required by small talk. Furthermore, interactions guided by specific prompts may feel scripted or unnatural, failing to capture the spontaneity and fluidity characteristic of genuine small talk.
%\textcolor{purple}{Ensuring the accuracy of an LLM's behavior requires more than just providing a specific system prompt, as demonstrated in our initial study. While a good system prompt sets the groundwork, potential biases in training data and the LLM's capacity to reflect user requests can result in drifts from the initial instructions. Without continuous monitoring, there is a risk of deviations from the system's intended behavior, especially as user behaviors and conversational trends evolve. This highlights the importance of regular adjustments to maintain accuracy and relevance throughout interactions.}

%The observer model is an instance of GPT plus a few methods for measuring how well a generated response meets each criterion of small talk. If GPT violates one of these criteria, the observer writes a new system prompt and sends it back to the model as feedback. As you said, it is continuously adapting the prompt. I can't hold it completely constant / on-script, but at least I can ensure the model self-corrects when it detects its own behavior is drifting. 

Hence, we introduce an \emph{observer model}, an instance of GPT that ``observes'' ongoing conversations, assessing whether responses from the ``speaking'' GPT model adhere to small talk criteria (Sec. \ref{sec:definitions}). If so, the generated response is relayed; otherwise, the observer generates a new system prompt and returns it to the speaking model as feedback. We call this technique \emph{feedback redirection}. As a result, the system self-corrects when it detects drifts in its behavior.

Rather than using the complete response of the speaking model as input to the observer, we utilize features defined based on the criteria outlined in Sec. \ref{sec:definitions}: brevity, tone, specificity, and coherence. The methods for their calculation are described below, followed by a description of the feedback prompts generated by the observer.

%If not, the observer generates a new system prompt and returns it as feedback. As a result, the model self-corrects when it detects drifts in its behavior. %A response is evaluated based on the criteria detailed in Sec. \ref{sec:definitions}: brevity, tone, specificity, and coherence.

%Therefore, we introduced an \emph{observer model}, which is includes a separate instance of GPT that ``observes'' an ongoing conversation and evaluates whether the generated responses of the main GPT model (the \emph{base model}) meet each criterion of small talk (Sec. \ref{sec:definitions}). When yes, the generated response is forwarded to the robot; otherwise, the observer generates a new system prompt and returns it as feedback to the base model. We call this technique \emph{feedback redirection}. As a result, the system self-corrects when it detects drifts in its behavior.

%Instead of using the full response of the base model as input to the observer model, we use features defined based on the criteria detailed in Sec. \ref{sec:definitions}: brevity, tone, specificity, and coherence. The ways how they are calculated are explained below, and the feedback prompts that the observer generates are described after that. 

\textbf{Brevity}. Setting a limit on the length of the generated responses enhances the practicality and user-friendliness of the small talk model, aligning with the natural flow of everyday conversations. To enforce this limit, the observer module defines an expected number of ``completion tokens''. Our iterative design process revealed that specifying a limit in words proved less accurate, as the number of words doesn't directly correspond to the number of tokens used in the model's internal representation \cite{openai_chatgpt}. This approach ensured more realistic and controlled conversations.

\textbf{Tone}. We employed the VADER model \cite{hutto2014vader} for sentiment analysis. The evaluation of tone and sentiment in a small talk response can be approached both per sentence and holistically. By combining both approaches, we gain a nuanced understanding of how the response contributes to the conversational tone, addressing both micro-level details and the macro-level coherence of the interaction. We estimated the relative weights of the holistic and per-sentence scores using the dataset collected in Sec. \ref{dataset}. A combined sentiment score (C) is calculated as follows:
\[
C = H \times w_H + \frac{1}{n} \sum_{i=1}^{n} s_i \times w_i
\]
In this formula:
\begin{align*}
& H \text{ is the overall score from VADER.} \\
& w_H \text{ is the weight assigned to the overall score. } \\
& n \text{ is the number of sentences.} \\
& s_i \text{ is the sentence-level score for the } i^{th} \text{ sentence.} \\
& w_i \text{ is the weight assigned to the } i^{th} \text{ sentence.}
\end{align*}
The score $C$ ranges from $-1$ to $+1$. A value between $-0.5$ and $0$ signifies a neutral response, and from $0$ to $1$ indicates positivity---both are acceptable for a small talk response. Responses with a score of $-0.75$ or lower are considered invalid by the observer module due to a strong negative tone. 

\textbf{Specificity}. The specificity of a response is assessed through NLTK's named entity chunker and part-of-speech tagging \cite{bird2009natural}. Counts of entities and descriptive words are normalized based on the maximum expected counts, derived from human responses in the dataset outlined in Sec. \ref{dataset}.

\textbf{Coherence}. To quantify coherence, we encoded each response into a sequence of tokens and derived embeddings using BERT \cite{devlin2018bert}. The calculated entropy of token embeddings of a response captures the uncertainty and diversity at each conversational turn. Subsequently, we gauged information gain by considering the entropy of the previous response and the weighted average of the entropies in the current response.

\textbf{Other Considerations}. As noted in Sec. \ref{dataset}, it is the nature of LLMs to offer assistance.
%GPT-3.5, though not explicitly an ``assistant'' model, the training data encompasses diverse internet text, including instances of assistant-like behavior \cite{wu2023brief}. Regardless, the model often exhibits inclinations to offer help or identify itself as an AI assistant. 
Yet, offers of help may result in conversations that sound too technical or formal. To mitigate this, the observer calculates the cosine similarity of embeddings to specified keywords of assistance, such as ``help'', ``assist'', and ``information''. We determined the list of specified keywords using the dataset collected in Sec. \ref{dataset}. %, guiding the conversation away from that of task-oriented assistance.

\subsection{Feedback Redirection}
%The observer serves a dual function, providing feedback to the model during the conversation and maintaining the load on the system. 
When the observer detects drifts in small talk characteristics, it provides feedback to the speaking model. \emph{Implicit feedback} allows the response but offers corrective guidance for unmet criteria. For instance, if the sentiment is excessively negative, a prompt might suggest, ``Your response was overly negative; aim for a neutral or lighthearted tone.'' Conversely, \emph{forced feedback} requires the speaking model to revise its response until full compliance is attained.
%When the observer detects drifts in small talk characteristics, it provides implicit or forced feedback to the speaking model. \emph{Implicit feedback} permits the generated response but gives corrective feedback to the speaking model describing unfulfilled criteria for the next request. For example, if the detected sentiment of a response is overly negative, a prompt such as, ``Your previous response was too negative. Remember to keep your responses neutral or lighthearted.'' In contrast, \emph{forced feedback} demands the speaking model regenerate its response until full compliance with the criteria is achieved. 
%The selection between forced feedback and implicit feedback by the observer module is governed by contextual factors and the quality of a given response. 
The observer opts for forced feedback when a response exhibits significant deviations along the measured criteria within the conversation. To facilitate timely responses, forced feedback is used sparingly as determined by a random factor, with a maximum limit of three regeneration attempts.

%\begin{center}
%\textcolor{purple}{\emph{Figure: Efficacy of Feedback Injections on Script Adherence}}
%\end{center}

%These feedback mechanisms encourage adherence to the initial system prompt throughout user interactions. These mechanisms also allow precise control over the load of the system to maintain an uninterrupted flow of dialogue while generating appropriate responses. 
The observer serves a dual function, providing feedback to the model during the conversation and maintaining the load on the system. The resulting model is evaluated on its ability to generate responses that align with small talk conventions.

\section{System Evaluation I: Chatbot Interactions}
The participants in the initial study engaged in 50 small-talk conversations with our observer model. The same experimental protocol and annotation guidelines for the initial study (Sec. \ref{dataset}) were used \cite{Ramnauth_2024}. Participants remained blind to the model they were interacting with and naive to the scope of the present research. A total of $50$ conversations with the observer model were transcribed, yielding $499$ responses with an average of $9.98$ responses per conversation ($SD = 0.14$). Of the $250$ generated responses, $106$ ($42.4\%$) responses were flagged by the observer with implied feedback, and $14$ ($5.6\%$) responses received forced feedback for a total of $23$ regeneration attempts ($M = 1.62$, $SD = 0.63$). 

We explored whether the observer's redirection was effective at improving the LLM's small-talk behavior. To compare the responses of ChatGPT-3.5 (base model) in the initial study (Sec. \ref{dataset}) to that with the observer model, we calculated the ``human-likeness'' of generated responses as described in Sec. \ref{ref:results-1} along the four small talk criteria. %and four conversational motives.

\begin{figure}[t]
    \centering
    \includegraphics[width=\columnwidth]{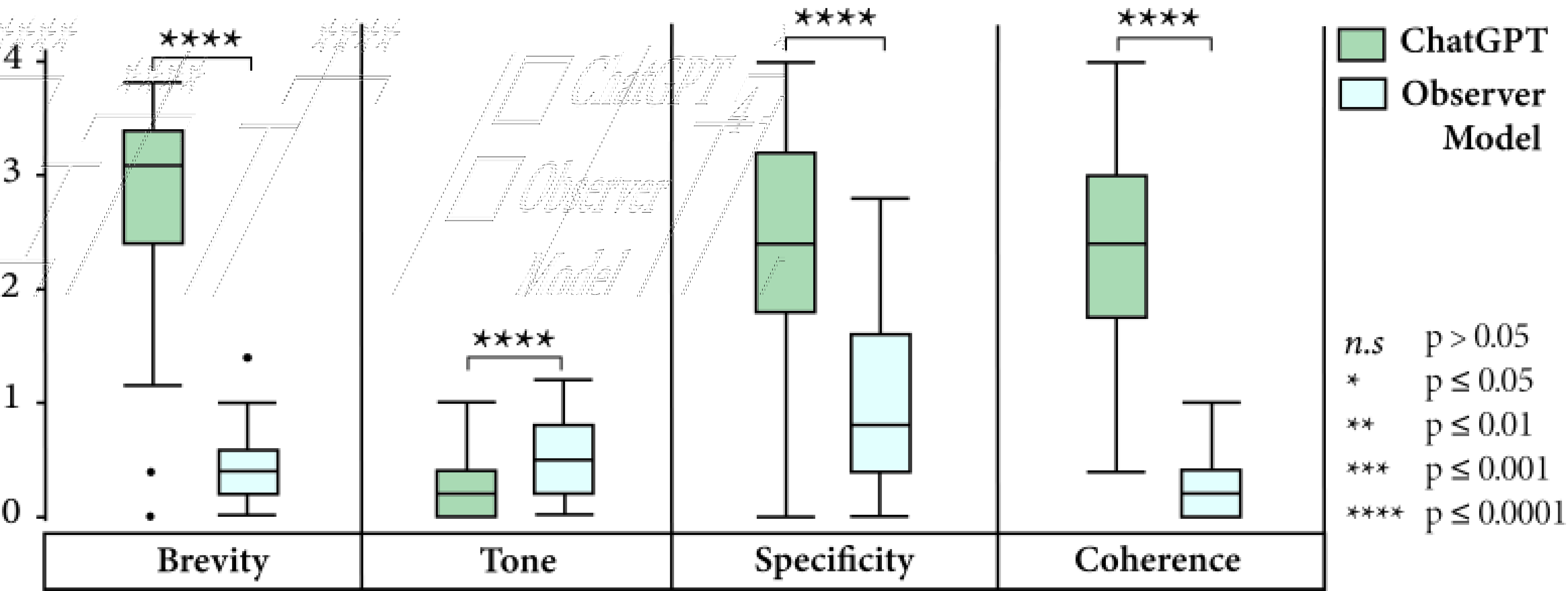}
    \caption{Human-Likeness of Observer v. Base Responses. The similarity of the models' small talk to that of the participants during text-based, chatbot interactions. Scores range from 0 (no difference) to 4 (highest absolute difference).}
    \label{fig:summary-observer}
\end{figure}

The Wilcoxon method with Holm-corrected significances indicates that the observer responses were significantly more human-like in that they were more concise ($Z = -8.17$, $p \leq 0.0001$), positive ($Z = 4.53$, $p \leq 0.0001$), less specific ($Z = -6.76$, $p \leq 0.0001$), and more thematically coherent ($Z = 4.53$, $p \leq 0.0001$) than the responses of the base system. 
%Observer-generated responses were also significantly less informative ($Z = -5.64$, $p \leq 0.0001$) and assistive ($Z = -8.26$, $p \leq 0.0001$), and significantly more expressive ($Z = -2.82$, $p = 0.0048$) and person-directed ($Z = 4.76$, $p \leq 0.0001$) than responses of the base system. 
Furthermore, a Brown-Forsythe test on the sum of differences across small-talk criteria indicates significantly less variability in human-likeness for the observer model than the base model ($F' = 15.47$, $p \leq 0.0001$). As summarized in Fig. \ref{fig:summary-observer}, the observer responses were more human-like across the criteria than the responses of the base model.

\section{System Evaluation II: Robot Interactions} 
A small-talk system should have the ability to engage effectively not only in virtual, text-based interactions but also in real-world, in-person scenarios. As a result, we applied the system to a robot to explore how well the system can navigate the nuances of face-to-face interactions. 

We used the robot Jibo \cite{Jibo} which stands 11 inches tall and has 3 full-revolute axes designed for 360-degree movement. Jibo's onboard capabilities allowed us to program personified behaviors such as naturalistic gaze, pose, and movement. We included a compact PC that communicates with the hardware and serves as local data storage. Additionally, we used a modular software architecture to allow for components of the small-talk system to be fully autonomous. The final system shown in Fig. \ref{fig:system} operates within the ROS framework.

\subsection{In-Person Evaluation}
A within-subjects case study was conducted where 25 volunteer participants, 15 men and 10 women, ages 19 to 45 ($M = 25.2$, $SD = 7.4$), interacted with the base and observer model for three conversations each. Each conversation spanned a minimum of eight turns, and the order in which participants interacted with the two models was randomized. This protocol yielded 150 conversations of 1725 responses in $\approx16.8$ hours of interaction, $40.5$ minutes ($SD = 10.2$) per participant. Following interactions with each model, participants provided open-ended feedback. We then conducted an informal thematic analysis and participant feedback was ultimately grouped into three primary themes. 

\emph{Response Content}. 21 participants expressed dissatisfaction with the base model's responses, noting its overly assistive and verbose tendencies, which led to conversations described as ``rambling'', ``dry'', and ``like speaking to a wall.'' This sentiment was echoed by $P_{25}$, who expressed frustration with the model's tendency to prioritize assistance over engaging in genuine conversation, stating, ``Even when I spoke about my own interests, it only cared about giving me help like I was a child always in need of help...'' On the other hand, in the observer condition, 23 participants remarked on how ``relevant,'' ``human-like,'' and ``natural'' were the robot's responses. For example, $P_{2}$ stated that the robot, ``engaged in small talk better than most of my friends would.'' 

\emph{Speech Delay}. Ten participants noted a delay in the robot's responses. As mentioned by $P_{7}$, ``natural, human-like speech has irregular pauses, ebbs, and flows,'' which can be difficult to predict or detect in real-time. The robot's speaking delay arises mainly from the processing time required for speech-to-text and text-to-speech, along with potential Wi-Fi latency. For the base condition, all five participants described the delay negatively (e.g., ``awkward'' and ``slow''), whereas all six participants described the delay positively (e.g., ``human-like'' and ``thoughtful'') for the observer condition.

\emph{Embodied Form}. 13 participants described the impact of the physical robot form on the quality of conversation. The feedback was mostly positive, highlighting that Jibo's ``animated'' and ``life-like'' movements made it ``more than a toy'' across conditions. Yet, three participants remarked on a lack of personality: ``[I]t's a bit misleading that it has a body and eyes and life-like movements but doesn't have a personality or experiences to share'' ($P_{14}$).

\subsection{Online Evaluation}
A limitation of our study thus far is the convenience sampling of mainly young adults from our local community. Our core research question is whether it is feasible to imbue robots with the capacity for small talk. While our findings indicate the feasibility and efficacy of such a system, we evaluate the system with a broader, more diverse demographic.  

To avoid experimenter bias, we randomly selected five participants from the in-person evaluation and their interaction video with each model. The videos were then edited to normalize speech delays using timestamps from the speech-to-text model. The resulting five video pairs were shared on Prolific \cite{palan2018prolific}, where online participants evaluated the robot in the base and observer conditions. The video presentation was randomized to mitigate order effects. Participants used 10-point Likert scales to rate the robot's human-likeness, naturalness, responsiveness, and casualness in each video.

\begin{figure}[t]
    \centering
    \includegraphics[width=\columnwidth]{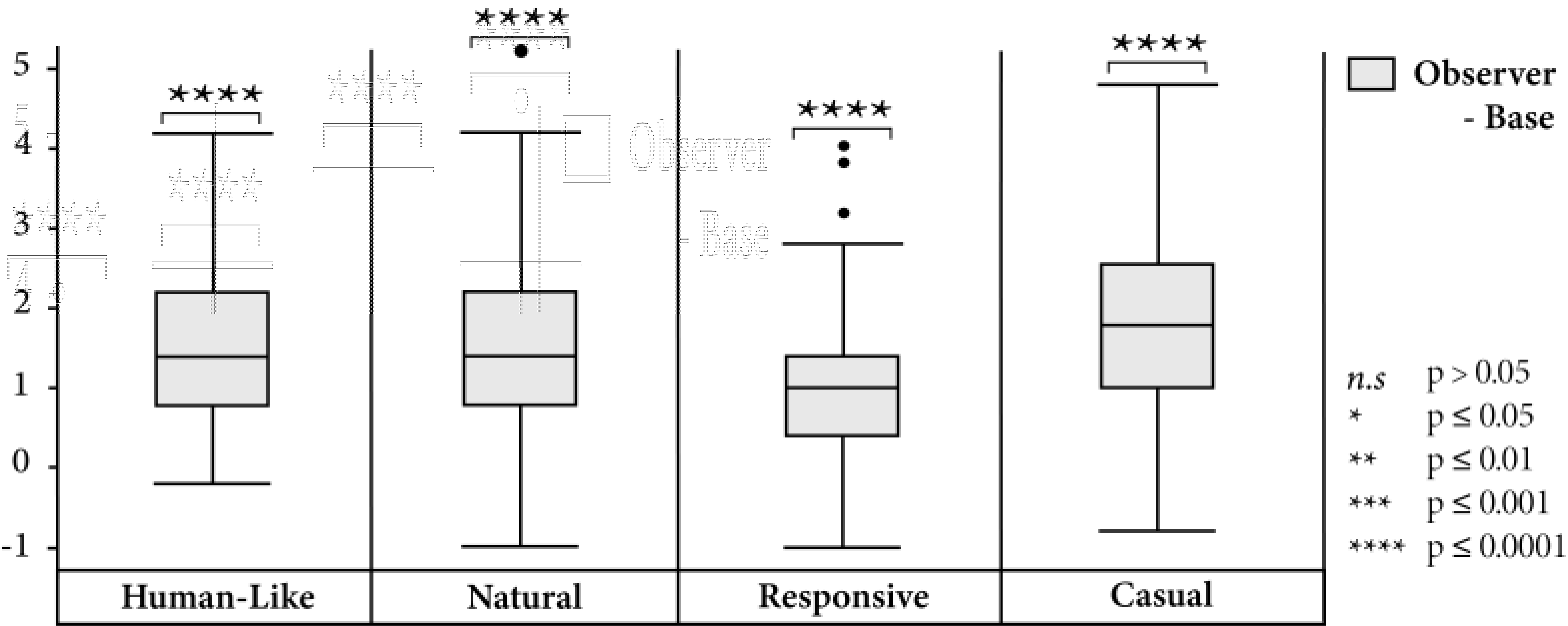}
    \caption{Observer v. Base in Online Assessments. Participant ratings of the human-likeness naturalness, responsiveness, and casualness of robot behaviors show that our system consistently outperformed the base model across all dimensions.}
    \label{fig:online}
\end{figure}

A total of 100 participants, 67 men and 33 women, ranging from ages 18 to 104 ($M = 43.1$, $SD = 18.7$) enrolled in the study. Assessments were averaged across the video pairs per participant and paired dependent t-tests were conducted between the two conditions. The difference in assessments between the observer and base model is denoted with $\Delta$. The resulting analysis indicates that observer behavior was more human-like ($t = 15.73$, $p \leq 0.0001$, $\Delta M = 1.53$, $\Delta SD = 0.10$), more natural ($t = 13.51$, $p \leq 0.0001$, $\Delta M = 1.50$, $\Delta SD = 0.11$), more responsive ($t = 11.22$, $p \leq 0.0001$, $\Delta M = 1.01$, $\Delta SD = 0.09$), and more like casual chat ($t = 15.80$, $p \leq 0.0001$, $\Delta M = 1.84$, $\Delta SD = 0.12$) than the behavior of the base model. 

In their open-ended feedback for each video pair, online participants echoed similar concerns as the in-person participants, such as the impact of specific response content and the robot's embodied form. For instance, $P_{61}$ remarks that the informative nature of the base system could be perceived as condescending: ``The first talk [base] contained a lot of stating facts or being somewhat snarky I found whereas the second [observer] was more of a casual conversation with someone that you haven't met before or seen in a long time.'' 

Despite video editing, there remains an irregular speaking delay due to the forced regeneration attempts made by the observer. Surprisingly, several participants stated that this perceptible delay added value to the interactions: ``The [base] robot emphasized its robotic AI form throughout the spoken exchanges. It... did not attempt to mimic human speech patterns or casualness like the first robot [observer]'' ($P_{87}$). 

Lastly, while most participants expressed a positive impact of the robot's form across conditions describing it as ``life-like'' and ``affirming'', a few participants voiced the opposite. For example, $P_{81}$ rated the base model higher in most video pairs because it is ``more like the thing it is supposed to be... an inanimate object that does not have feelings.'' %In all, these evaluations yielded several nuanced considerations for the future of small-talk capable systems. 

\section{Discussion}
%It is against the nature and design of LLMs to engage in the free-flowing, casual dialogue that is small talk. Despite advancements in natural language processing, appropriate and human-like small-talk conversations remain a challenge for LLMs. 
Though LLMs show substantial potential in enabling natural language capabilities for social robots, achieving seamless and contextually appropriate casual dialogue remains a challenge. We began by assessing the capacity of current LLMs to participate in small talk, identifying key areas for improvement. Subsequently, we presented a novel method of feedback redirection to ensure LLM-generated responses align with small talk conventions. Through three evaluations, we examined the system's efficacy in sustaining autonomous small-talk interactions.

First, participants engaged in text-based chat interactions, where our system outperformed a baseline LLM with the same initial prompt. Next, we explored whether this success continues for novel, in-person human-robot conversations. We not only showed the system's robustness in a real-world setting but also the inadequacy of an ``out-of-the-box'' LLM for such interactions. Lastly, online assessments of the robot interactions affirmed that our system resulted in a robot significantly more capable of natural, human-like dialogues.

%Although we implemented the system with a single LLM for a particular robot platform, the system relies on a method that can easily generalize to other platforms and LLMs. 
While the design and internal representation of different LLMs and robot platforms will have various requirements, the concept of feedback redirection to ensure generated responses adhere to an initial system prompt is not unique to ChatGPT 3.5 or Jibo. Further, it may generalize beyond small-talk behavior to other actions and domains. Future research should explore how generalizable and effective feedback redirection is to system prompt adherence for various LLMs, platforms, and behaviors. 

In conclusion, our study offers specific technical contributions to the field of conversational AI and social robotics. Our novel system enables robots to engage in authentic and contextually appropriate small talk autonomously. By addressing a crucial gap in current LLM use and ability, our study offers practical insights and a tangible framework for more natural and engaging human-robot interactions.

%In all, small talk is a counterpoint to the inherent nature of LLMs to provide task-oriented assistance and information. As a result, this paper provides a case study of how feedback redirection can help build more robust and social conversational agents. 

%A limitation of our study is the convenience sampling of participants from our local community. Our core research question is whether it is feasible to imbue a robot with the capacity to engage in naturalistic small-talk conversations with users. While our findings indicate the feasibility of such system development, future research could explore the expectations and interactions of a broader and more diverse user demographic regarding small-talk-capable systems.  

\section*{Acknowledgments}
\noindent This work was partially funded by the National Science Foundation (NSF) under grants 1955653 and 2106690, the Office of Naval Research (ONR) grant N00014-24-1-2124, and the JST Moonshot R\&D under grant JPMJMS2011, Japan. Rebecca Ramnauth is supported by the NSF GRFP and the NASEM Ford Predoctoral Fellowship.

%\bibliographystyle{IEEEtran}
%\bibliography{references.bib}


\end{document}